\documentclass[12pt]{article}
\pdfoutput=1

\usepackage[utf8]{inputenc}
\usepackage{graphicx}
\usepackage{subcaption}
\usepackage{amsmath}
\usepackage{amssymb}
\usepackage{geometry}
\usepackage {hyperref}
\usepackage {abstract}
\title{QKVA grid: Attention in Image Perspective\\ and\\ Stacked DETR}
\author{Sheng Wenyuan}
\date{July 2022}

\begin{document}

\maketitle

\begin{abstract}
We present a new model named Stacked-DETR(SDETR), which inherits the main ideas in canonical DETR\cite{2020detr}. We improve DETR in two directions: simplifying the cost of training and introducing the stacked architecture to enhance the performance.
To the former, we focus on the inside of the Attention block and propose the QKVA grid, a new perspective to describe the process of attention. By this, we can step further on how Attention works for image problems and the effect of multi-head. These two ideas contribute the design of single-head encoder-layer.
To the latter, SDETR reaches better performance(+0.6$AP$, +2.7$AP_S$) to DETR. Especially to the performance on small objects, SDETR achieves better results to the optimized Faster R-CNN baseline, which was a shortcoming in DETR.
Our changes are based on the code of DETR. Training code and pretrained models are available at \url{https://github.com/shengwenyuan/sdetr}.
\end{abstract}

\section{Introduction}
The problem of object detection includes both the classification and the location of the targets. In conventional ways, the one-step framework like YOLOv3\cite{2018yolov3} addresses it by setting anchor boxes to preload the base location and predefine the approximate boxes. To raise the collection from wider field, it uses FPN architecture. YOLOv4\cite{2020yolov4} even enhances it wider by adapting PAN and SPP designs. The two-step framework like Faster RCNN\cite{2015frcnn} uses region proposal in RPN layer. \\
The wider field is explainable by making an analogy with human observation\*. People's recognition of different objects requires global information: in some cases, we could not classify the objects without the help of wider environments. Transformer is based solely on attention mechanisms, which would be useful beyond the convolution blocks.
Vision Transformer(ViT\cite{2020vit}) reaches almost the same accuracy compared with traditional frameworks, which treat the vision pixels as language tokens. Convolutional vision Transformer(CvT\cite{2021cvt}) combines attention layer with convolutional projection.
We inherit the idea of End-to-End object detection(DETR), which connects the CNN backbone and Transformer module in sequence and turns the result of decoder into a prediction list directly by using Hungarian match to calculate the losses.

\section{Background}
\subsection{Transformer's Global Vision}
\begin{figure}[h]
  \begin{minipage}[b]{0.35\textwidth}
    \centering
	\includegraphics[width=0.8\linewidth]{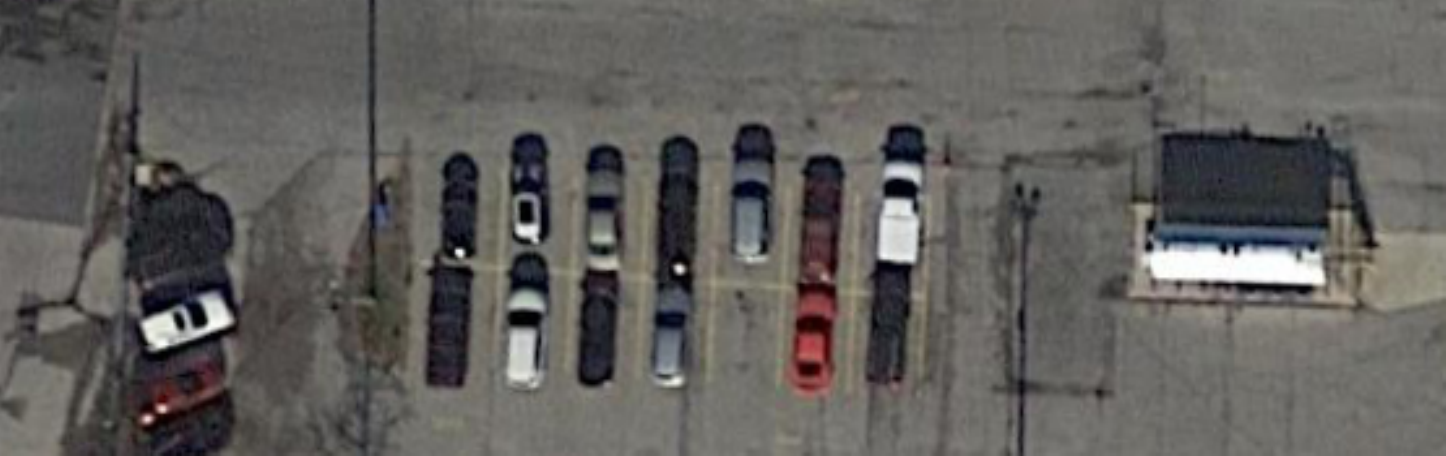}
	\subcaption{The white roof looks like a van.}
    \label{fig:car1}

	\includegraphics[width=0.3\linewidth]{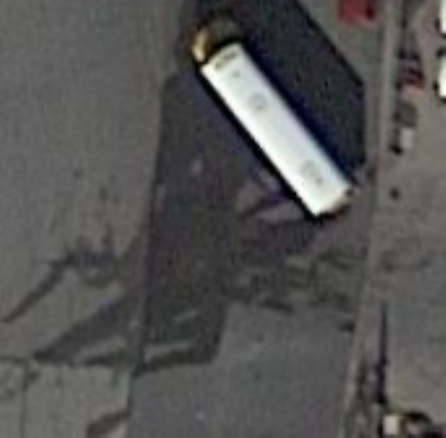}
	\subcaption{The bus looks like a white roof.}
    \label{fig:car2}
  
  \end{minipage}
  \begin{minipage}[b]{0.6\textwidth}
    \centering
    \includegraphics[width=0.8\linewidth]{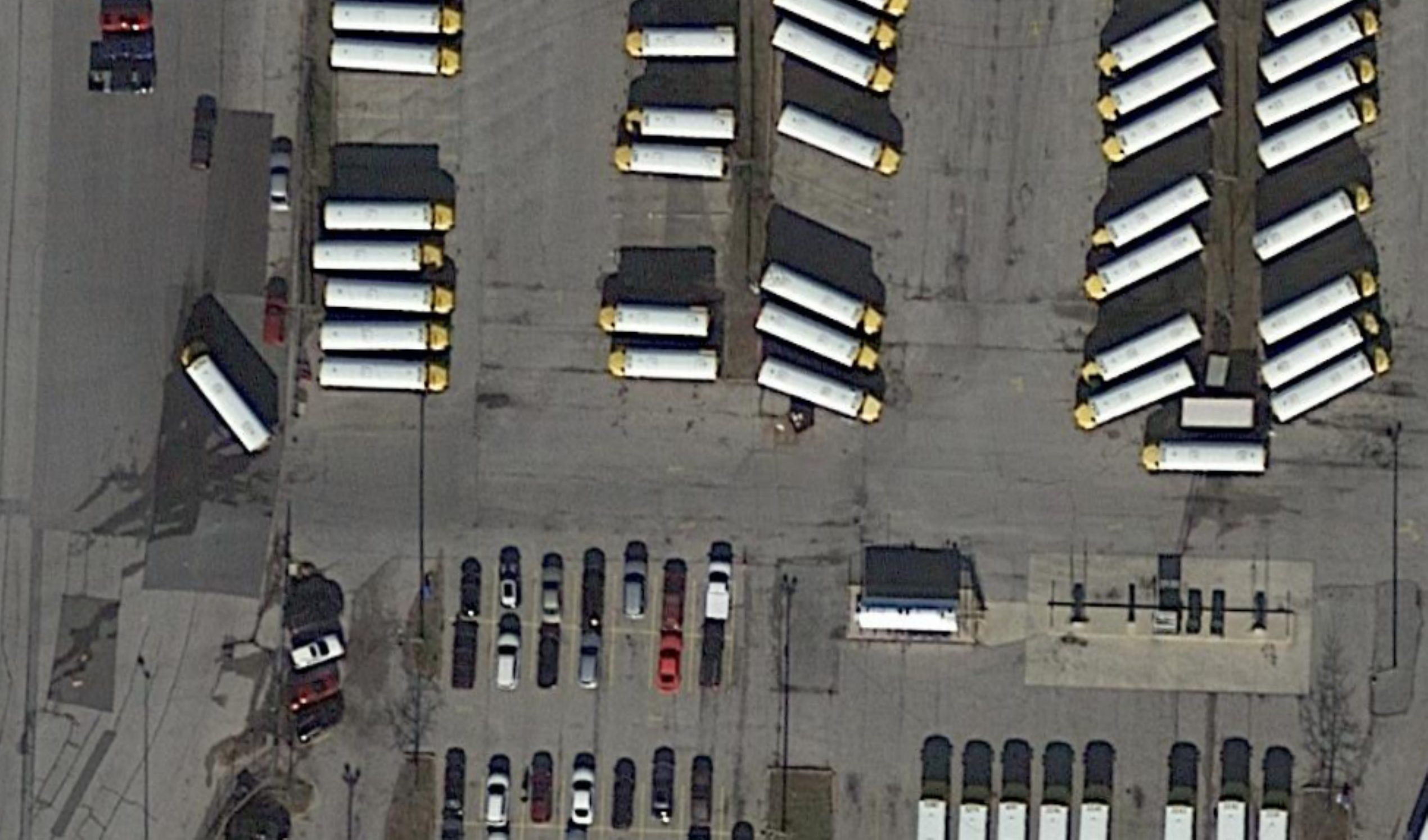}
	\subcaption{If we got the whole image, the answer is clear.}
  \end{minipage}
  \caption{}
  \label{fig:cars}
\end{figure}

The conventional anchor based module is hard to train the qualified accuracy merely from local inputs. Fig\ref{fig:cars} depicts one specific condition that the local information could never provide accurate direction because their judgments are dependent on a much wider environment, like the whole image.
For example, if a car is classified with high accuracy, it should be helpful to classify other similar rectangles around it and preclude some confusing choices.

\begin{figure}[ht]
  \centering
   \includegraphics[width=0.4\linewidth]{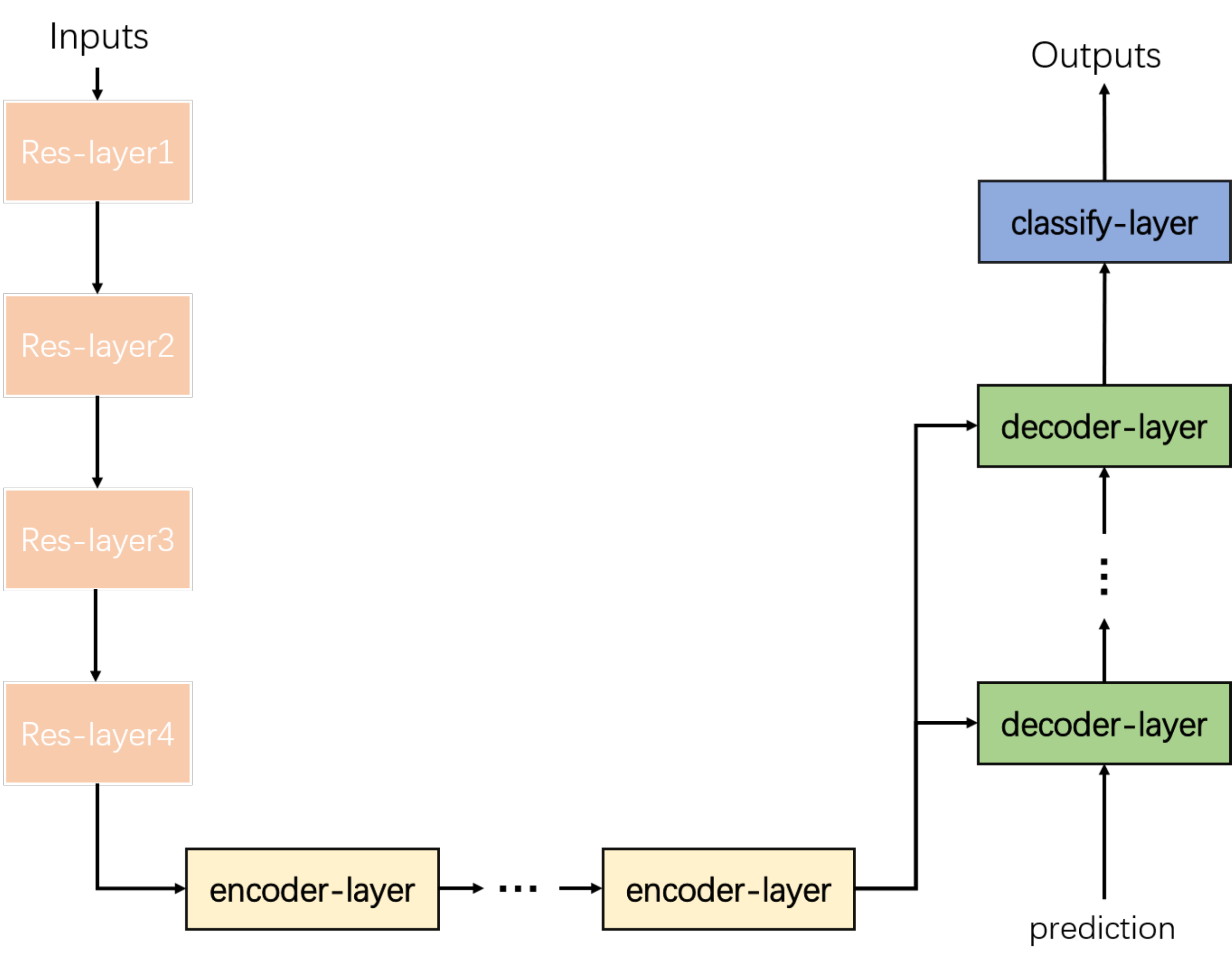}
   \caption{}
   \label{fig:detr1}
\end{figure}

\subsection{DETR Architecture and Human Observation}
\begin{figure}[h]
  \centering
   \includegraphics[width=0.5\linewidth]{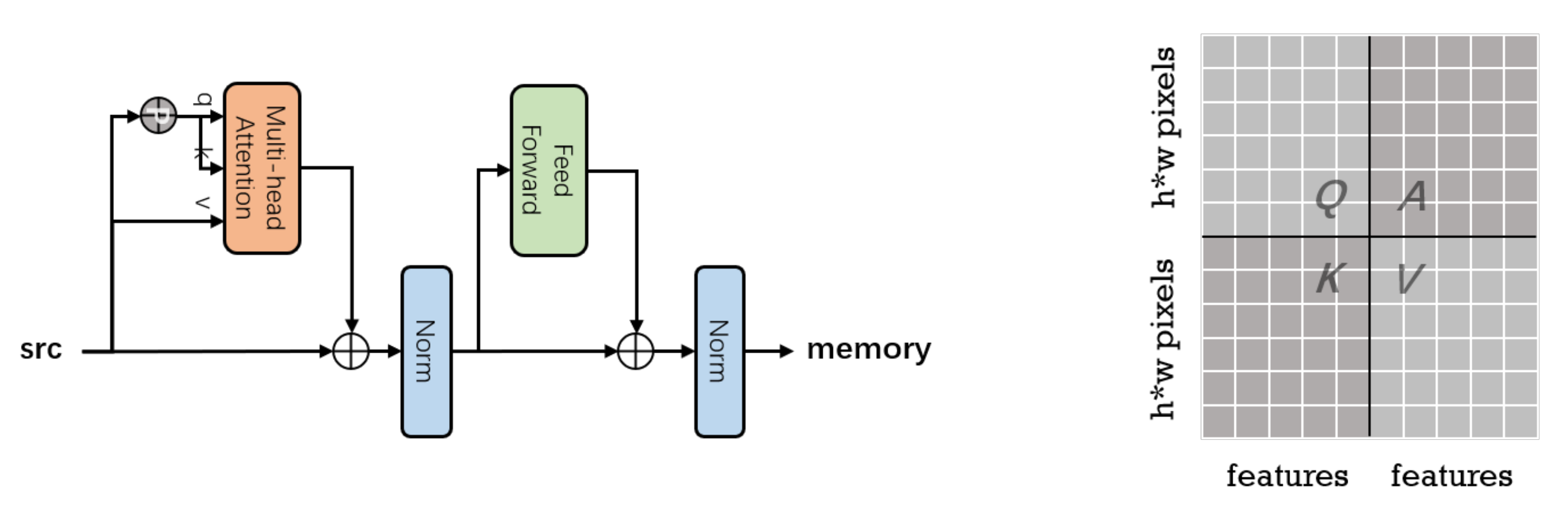}
   \caption{}
   \label{fig:detr2}
\end{figure}

\begin{figure}[htbp]
  \centering
   \includegraphics[width=0.45\linewidth]{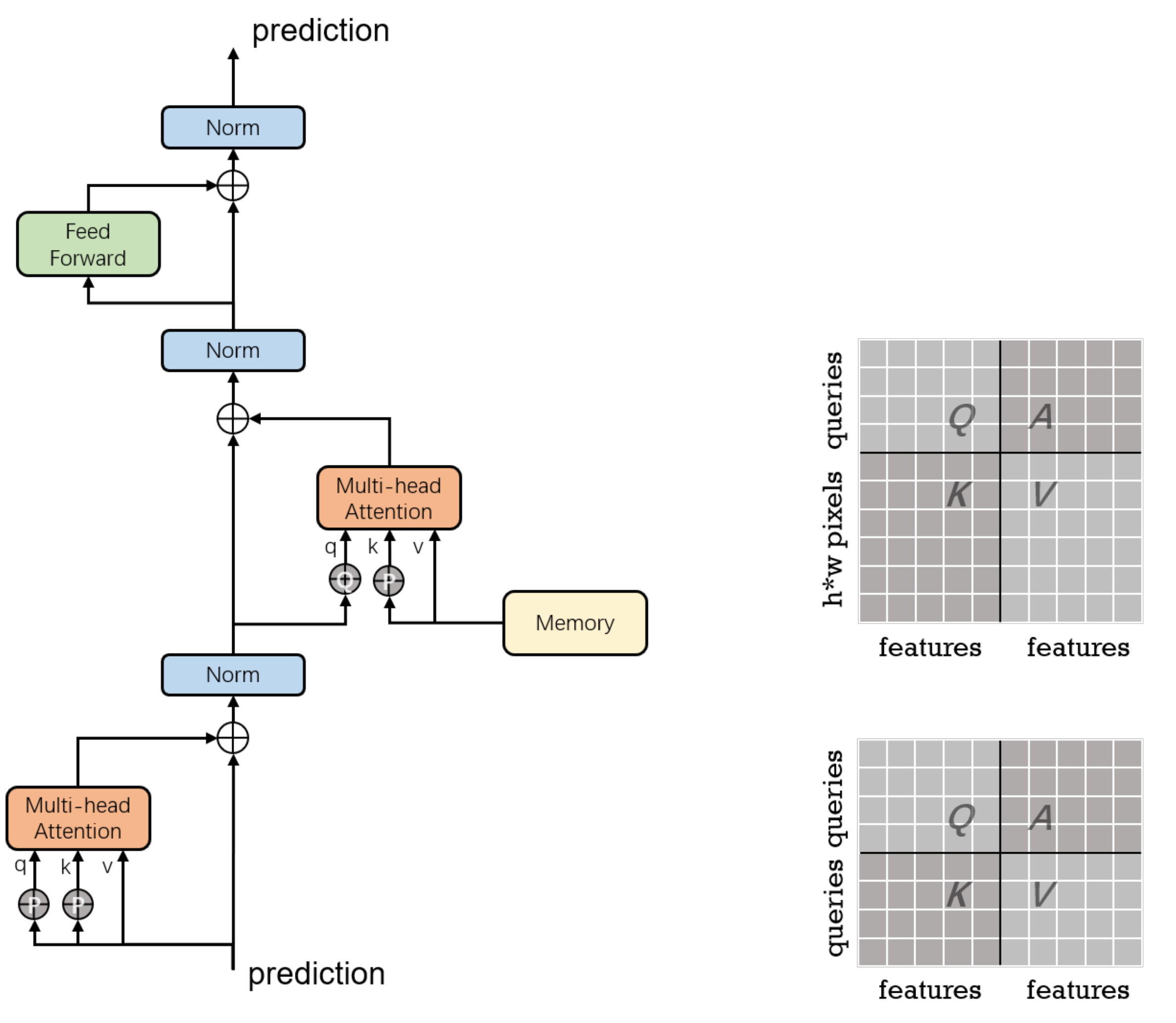}
   \caption{}
   \label{fig:detr3}
\end{figure}

DETR uses 4 layer ResNet backbone to learn the local information from input images. Added with positional encoding, the abstract images become the input of Transformer's encoder-layer.
DETR designs a single pipeline to connect many layers of encoders and decoders in sequence, which keeps all the architecture of the canonical Transformer module. We redraw the content of an encoder and a decoder with central nexus in Fig\ref{fig:detr1}. Though the implementation still works the same with DETR, we reschedule the layout of blocks in another perspective: an analogy to human observation.
In an encoder-layer (see in Fig\ref{fig:detr2}, left), a self-attention block collects orthogonal units of input on the image dimension, and a feed-forward block, which contains two full-connected layers on the feature dimension, collects full information.
In a decoder-layer (see in Fig\ref{fig:detr3}, left), we centralize the prediction line. At the first layer, the inputs are all zero and all the parameters work for training these inputs into the answers. 
The analogy starts from people's general judgment of the topic of the image. So in any decoder-layer, the prediction is contributed by a cross-attention block that collects global information from abstracted pictures (also named memory). 
Then people focus on certain targets to determine classifications and locations, which maps the feed-forward block.
Finally, people have to check if their judgments follow the real logic in living. Hence at the start of the next decoder-layer, the prediction matrix needs to absorb the result of a self-attention block, which collects more global information to the mainline. People repeat all these actions to get a more accurate result and it explains why setting multi-layers of decoder is essential to the architecture.


\section{Related work}
\subsection{Stacked DETR}
\begin{figure}[h]
  \centering
   \includegraphics[width=0.4\linewidth]{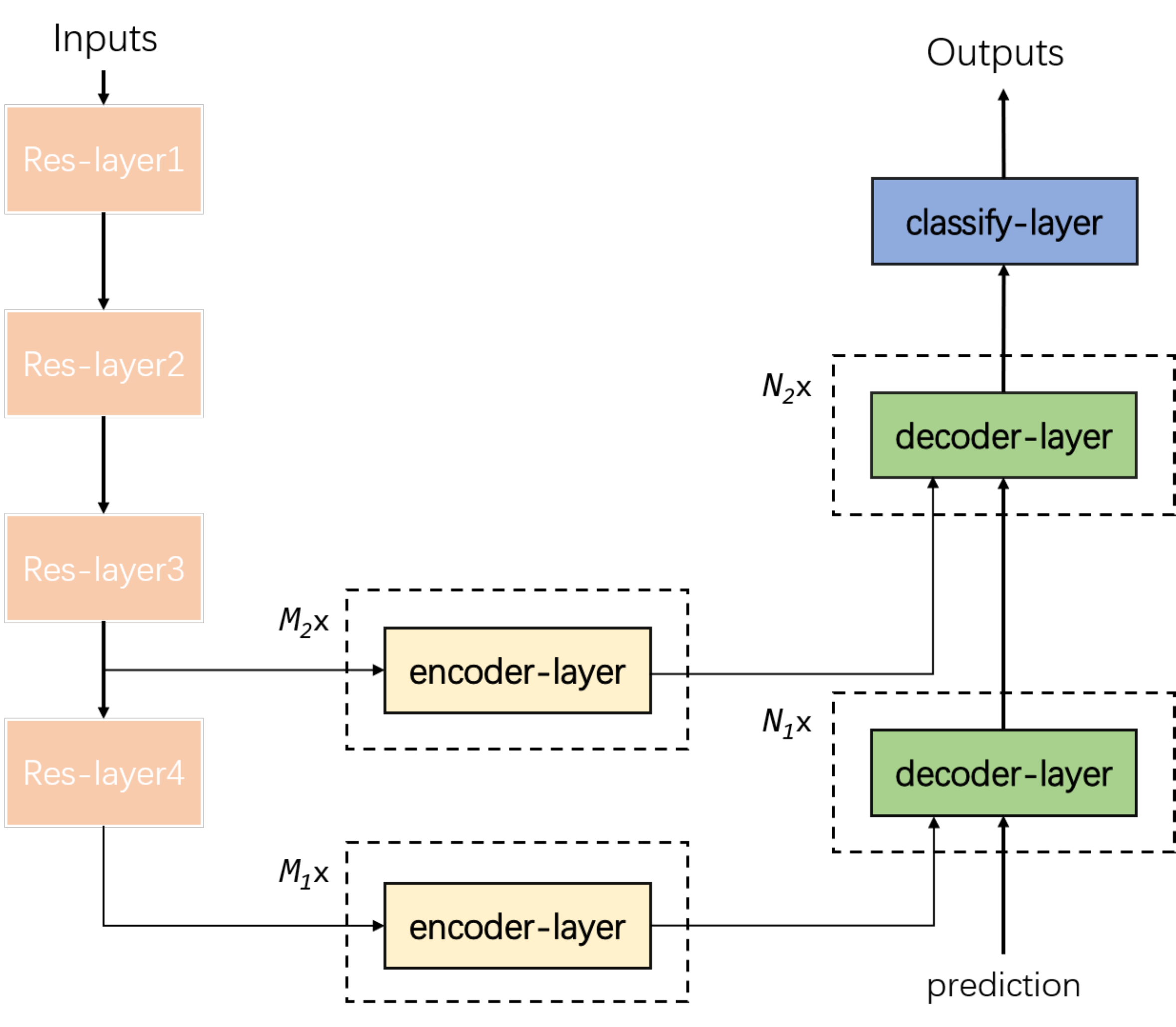}
   \caption{}
   \label{fig:stack1}
\end{figure}

Based on the above understanding of DETR's working mechanism, we can introduce our little enhancement now. Similarly, our point starts from human observation: it is not completed in one step nor on one scale. If one could discern a more accurate location by focusing on outlines, the module should use relatively less-abstract and detailed images as well.
Our stacked DETR \ref{fig:stack1} rearranges the correlations between ResNet\cite{2015resnet} backbone, encoder-layer, and decoder-layer. We add one transformer module right after the third layer of ResNet: the outputs of each encoder line contribute in parallel and two packs of decoder-layers are connected in sequence.
Though the idea of building `stacked' architecture deserves more than two levels, we are forced to limit the architecture's scale because of computational cost. We leave the detailed reasons in the Experiments part.
%
%To be continue: add pictures about remote cars

\subsection{Various Heads}
In the Appendix, we analyze one character of using multi-heads: consuming n-times of memory occupation. For instance, if we set the size of Res-layer4's output as $h*w$, the Res-layer3's output would have $2h*2w$, which leads $(2h*2w)^2$ amount of memory occupation in one encoder block. The model makes $((2h*2w)^2)*n$ in total heads. By using single-head attention, term $n$ is canceled. \\
However, our test shows that there is performance loss when the model uses single-head Attention in both encoder-layers and decoder-layers. Moreover, the decoder-layer relies on multi-head to separate two kinds of predictions: classification and box location. To compensate for the extra memory occupation from our stacked pipeline, it is necessary to implement the single-head Attention in the maximum size layers. We finally set the single-head attention in Res-layer3's encoder-layers and keep the multi-head attention in every other layer.

\section{Experiments}
\subsection{Resources}
\textbf{Computational amount.} 
It seems better to add more levels of transformers by encoding in parallel and decoding in sequence. However, layers at a higher level have many more computational variables to be computed in attention. In the Appendix, we prove that $computation = h*h'*(w+w')$. ResNet is four times the length of $h$ between levels, which leads to sixteen times of computation. It forces us to balance the accuracy and the scale of computational amount. \\
\textbf{Memory occupation of GPU.}
When training any version of model based on Transformer, we observe that the memory occupation of GPU varies greatly and the maximum may surpass the GPU’s hardware upper limit, which incurs breaking. The dot-product result of matrix-Q and matrix-K is a square with dimensions $h*w$, the scale of abstracted pictures. Since GPU has to save the square matrix for the next dot-product, it consumes $(h*w)^2$ in extra. Tab\ref{tab:training2} shows that single-head is useful to reduce the memory occupation. \\
Memory occupation of GPU has always been a bottleneck in our tests. To fully train a standard DETR-related model, the cost is equal to one month of living expenses of a junior student. It forces our model must workout under cheaper resources and shorter time. On this basis, we develop tiny-scaled inputs, which could verify our ideas by using smaller configurations.

\begin{table}[h]
  \centering
  \renewcommand{\arraystretch}{1.5}
  \begin{tabular}[t]{c|c|c|c|c}
    \hline
    Model(layers + heads)             & Batch=2    & Batch=4  & Batch=8  & Batch=16\\
    \hline
    Fixed DETR(6-6 + 8-8)                   &  9601 & 16795 & 30989 & 59739\\
    Fixed DETR(6-6 + 1-8)                   &  6515 &	10359 & 17181 & 28171\\
    Fixed Stacked-DETR(3-3-3-3 + 8-8-8-8)	& 44199 &	overflow & overflow &	overflow\\
    Fixed Stacked-DETR(3-3-3-3 + 8-8-1-8)	& 10761 &	19091 &	35765 &	overflow\\
    \hline
  \end{tabular}
  \caption{The unit is MiB. `Fixed' means the inputs are set in $1333*1333$ pixels to trace the stable memory occupations. LAYERS has the structure of `M1-N1-M2-N2'. HEADS describes the head number in each layer, which follows the sequence as `level4's encoder - level4's decoder - level3's encoder - level3's decoder'.}
  \label{tab:training2}
\end{table}

\begin{table}[h]
  \centering
  \renewcommand{\arraystretch}{1.5}
  \begin{tabular}[t]{c|c|c|c|c}
    \hline
    Model(layers)   & $AP$    & $AP_S$  & $AP_M$  & $AP_L$\\
    \hline
    6-6-0-0 & 28.0 &	8.5 &	28.7 &	48.3\\
	\hline
    5-5-1-1 & 28.5 &	8.2 &	30.1 &	48.9\\
    4-4-2-2 & 29.1 &	8.9 &	30.2 &	50.3\\
    3-3-3-3 & 29.7 &	9.2 &	31.6 &	51.0\\
    2-2-4-4 & 29.7 &	9.4 &	31.8 &	49.3\\
    4-4-0-2 & 27.7 &	7.3 &	28.2 &	48.6\\
    \hline
  \end{tabular}
  \caption{We test five models with tiny-scaled input, which is the fixed $512*512$ pixels. Their layers follow the sequence as `M1-N1-M2-N2'. In this format, DETR has `6-6-0-0' layers.}
  \label{tab:training1}
\end{table}

\begin{figure}[h]
  \centering
   \includegraphics[width=1.0\linewidth]{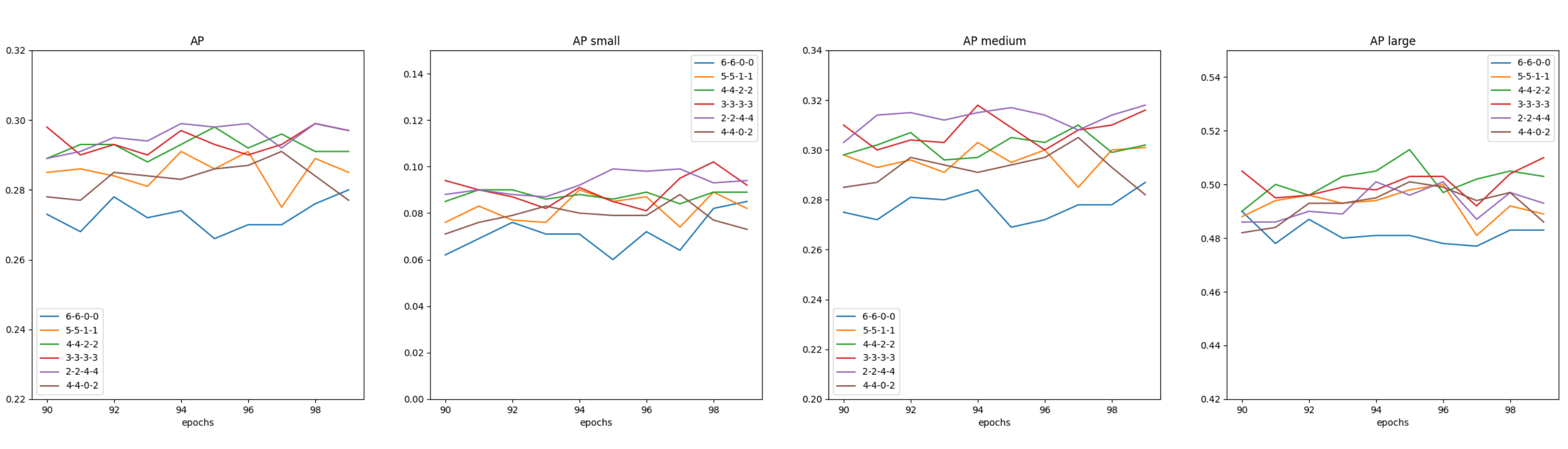}
   \caption{We record all six models' performances of the last ten epochs. It shows that stacking architecture outperforms DETR model.}
   \label{fig:tt}
\end{figure}

\subsection{Tiny-scaled Inputs}
Though DETR significantly outperforms competitive baselines, it needs to be trained for a long time with multi tough GPUs. By delimiting the same size of every input, both the scale of computational amount and memory occupation are greatly reduced. Tab \ref{tab:training4} proves that there is a positive correlation between the tiny-scaled version and the full-scaled inputs version.\\
To preclude the positive correlation between the number of layers and the performance, we select the DETR module with six encoder-layers and six decoder-layers as our control group. In other words, both M1 plus N1 and M2 plus N2 should be equal to six (see Fig\ref{fig:stack1}). We used an RTX-A4000 GPU and trained 100 epochs. Tab \ref{tab:training1} provides that our stacked model raises the average precision by almost 2\%. Almost all three sizes of objects got raised as well.

\begin{table}[h]
  \centering
  \renewcommand{\arraystretch}{1.5}
  \begin{tabular}[t]{c|c|c|c|c}
    \hline
    Model(layers) & $AP$  & $AP_S$  & $AP_M$  & $AP_L$\\
    \hline
	6-6-0-0 & 39.5 & 17.5 &43.0 & 59.1\\
	4-4-0-2 & 39.6 & 18.3 &43.0 & 59.0\\
    \hline
  \end{tabular}
  \caption{We test two models with FULL-scaled input to have min size 800 and max size 1333. Each model has trained 150 epochs. In fig\ref{fig:tt}, `4-4-0-2'(brown) has better performance than `6-6-0-0'(DETR, blue). Here the full pair keeps the correlation, which proves the validity of our tiny-scaled tactic.}
  \label{tab:training4}
\end{table}

\subsection{Single-head}
Though replacing multi-head into single-head saves the memory occupation, the performance consumes. However, with the help of tiny-scaled testing, the results(see in Tab \ref{tab:training5}) show that when the model only uses single-head in the ResNet3's encoder layer, the performance is approximately the same.\\
\begin{table}[h]
  \centering
  \renewcommand{\arraystretch}{1.5}
  \begin{tabular}[t]{c|c|c|c|c}
    \hline
    Model(layers + heads) & $AP$  & $AP_S$  & $AP_M$  & $AP_L$\\
    \hline
	3-3-3-3 + 8-8-8-8     & 29.7 & 9.2 & 31.6 & 51.0\\
	3-3-3-3 + 8-8-1-8     & 29.8 & 9.8 & 30.9 & 50.4\\
    \hline
  \end{tabular}
  \caption{The experiment on single-head attention}
  \label{tab:training5}
\end{table}

To directly show our understanding of multi-head, we designed two pairs of tests running on an A100 GPU (see tab \ref{tab:training2}).  With the growth of batch size, the memory occupations almost increase by double. Especially to our stacked DETR, its memory occupation overflows even at 4 batch size. By implementing single-head on encoder-layers, our memory cost decreased by many folds.

\subsection{Our scores}
We train a full-scaled inputs version in 150 epochs, which can be compared with one of canonical DETR's evaluation \footnotemark.\footnotetext{https://gist.github.com/szagoruyko/b4c3b2c3627294fc369b899987385a3f}
Our model \footnotemark is implemented basing on the code of DETR. \footnotetext{https://github.com/shengwenyuan/sdetr}
It retains every hyperparameter, including learning rate, augmentation, dropout ratio, etc. Tab \ref{tab:training3} shows that stacked-DETR achieves higher score in $AP$(+1.1), which is led with $AP_M$(+1.1) and further improvement in $AP_S$(+3.4).

\begin{table}[h]
  \centering
  \renewcommand{\arraystretch}{1.5}
  \begin{tabular}[t]{c|c|c|c|c|c}
    \hline
    Model(layers + heads)           & \# params & $AP$    & $AP_S$  & $AP_M$ & $AP_L$\\
    \hline
    DETR(6-6-0-0 + 8-8-0-0)         & 41.3M     &39.5    & 17.5   &43.0   &59.1 \\
    Ours epoch147(3-3-3-3 + 8-8-1-8)& 41.5M     &40.6    & 20.9   &44.1   &58.1 \\
    Ours epoch150(3-3-3-3 + 8-8-1-8)& 41.5M     &40.5    & 20.2   &44.2   &58.0 \\
    \hline
  \end{tabular}
  \caption{Same as before, LAYERS has the structure of `M1-N1-M2-N2'. HEADS describes the head number in each layer, and they follow the sequence as `level4's encoder - level4's decoder - level3's encoder - level3's decoder'.
Our training environment is four RTX-3090 GPU with 24GB memory and Intel Xeon CPU. To one epoch, the canonical DETR takes around 53 minutes for training and 103 seconds for testing. Our model takes around 56 minutes for training and 112 seconds for testing. To the GPU memory usage, the canonical DETR takes 20GB and our model takes 22GB. In short, the extra cost of training is less than 10\%.}
  \label{tab:training3}
\end{table}

We also train in 500 epochs, which makes $AP$(+0.6), $AP_M$(+0.4), and $AP_S$(+2.7). Stacked-DETR has lower performance in $AP_L$(-0.9).
\begin{table}[h]
  \centering
  \renewcommand{\arraystretch}{1.5}
  \begin{tabular}[t]{c|c|c|c|c|c}
    \hline
    Model(layers + heads)           & \# params & $AP$    & $AP_S$  & $AP_M$ & $AP_L$\\
    \hline
    DETR(6-6-0-0 + 8-8-0-0)         & 41.3M     &42.0    & 20.5   &45.8   &61.1 \\
    Ours(3-3-3-3 + 8-8-1-8)         & 41.5M     &42.6    & 23.2   &46.2   &59.2 \\
    \hline
  \end{tabular}
  \label{tab:ours500}
\end{table}

\section{Conclusion}
Our changes include the multi-levels of Attention pipelines in model, the single-head encoder-layer in one Attention block, and the scale of inputs to train with a simpler resource. SDETR shows the potential of being a lite, end-to-end model: each part can be designed separately to balance resources and performances.
SDETR outperforms on both large and small objects than Faster R-CNN baseline on COCO dataset. It concretes the availability of end-to-end object detection with transformer. \\
The resource has been the bottleneck during our whole research. We will start deeper research on Attention, like whether the effect of multi-head is worth those extra memory resources.\\

\bibliography{v3}

\clearpage
\section*{Appendix}
\textbf{A. Image Attention}\\
Proposed by Ashish Vaswani et al.\cite{2017attention}, the transformer model is becoming a more popular resolution to vision tasks. However, though the Transformer module which contains the scaled dot-product attention is widely used in many ideas\cite{2020vit} \cite{2021cvt}, people are used to treating the Transformer as a unit block and do not step further to its inside detail. Hence it is needed to explain why the scaled dot-product attention works well beyond natural language processes. 
Here we propose a new view to explain the task of Q, K, V matrices. If we set a hidden matrix P to separate the formula as $P=Q\cdot K^T$ and $A=P\cdot V$, the traversing process will be traced as follows.

\begin{figure}[h]
  \centering
  \begin{minipage}[t]{0.65\textwidth}
  \centering
  \includegraphics[width=0.9\textwidth]{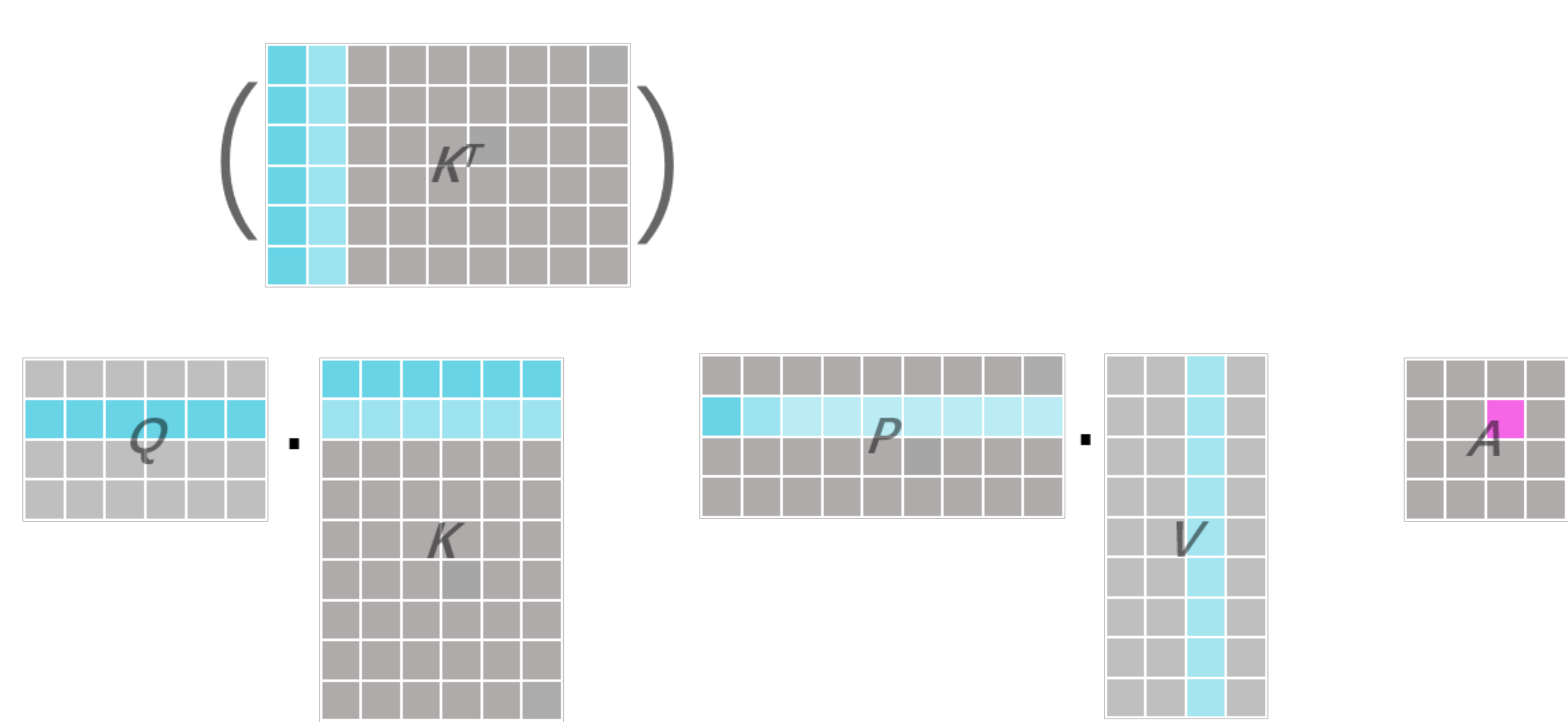}
  \caption{$Q\cdot K^T \cdot V$}
  \label{fig:att1}
  \end{minipage}
  \begin{minipage}[t]{0.3\textwidth} 
  \centering  
  \includegraphics[width=0.6\textwidth]{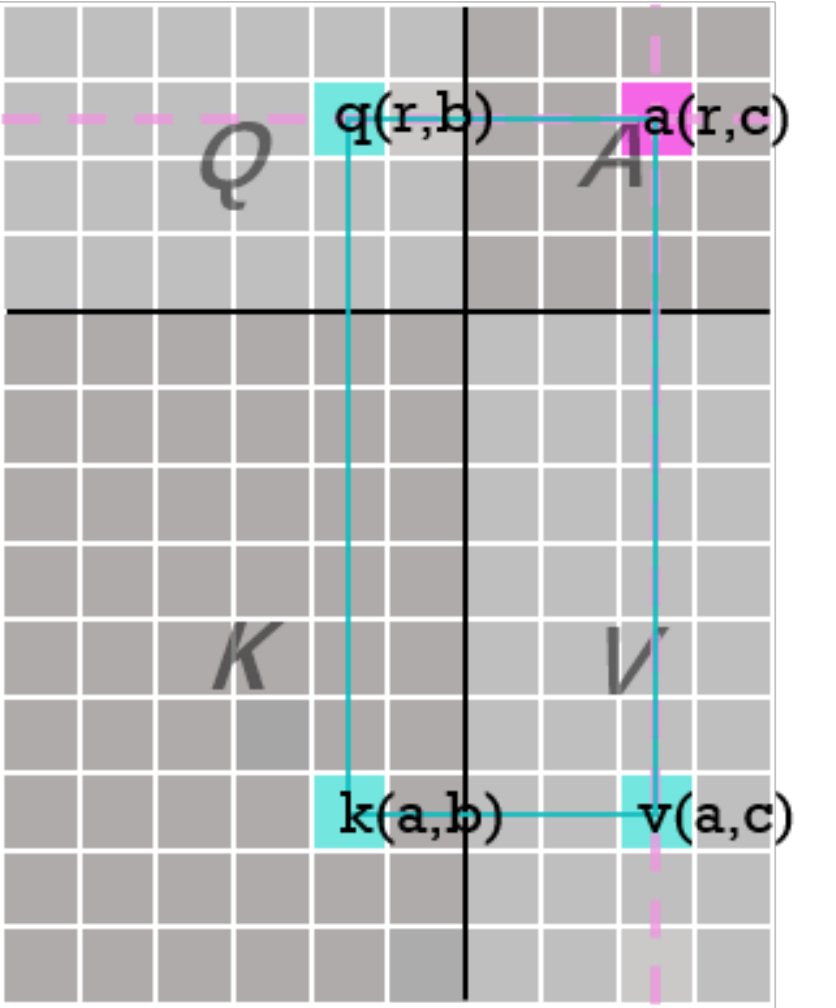}
  \caption{QKVA grid}
  \label{fig:att2}
  \end{minipage}   
\end{figure}

\begin{align}
P_{(a,b)}
=&Q_{(a,1)}\cdot K^T_{(1,b)} + Q_{(a,2)}\cdot K^T_{(2,b)} + \dots + Q_{(a,w)}\cdot K^T_{(w,b)} \\
=&Q_{(a,1)}\cdot K_{(b,1)} + Q_{(a,2)}\cdot K_{(b,2)} + \dots + Q_{(a,w)}\cdot K_{(b,w)} \notag
\end{align}
\begin{align}
Attention_{(x,y)} 
=&\quad   P_{(x,1)}\cdot V_{(1,y)} +              P_{(x,2)}\cdot V_{(2,y)} + \ldots \ldots +   {}           P_{(x,w)}\cdot V_{(w,y)} \\
=&\quad   Q_{(x,1)}\cdot  K_{(1,1)}\cdot V_{(1,y)} + Q_{(x,2)}\cdot K_{(1,2)}\cdot V_{(1,y)} + \ldots + Q_{(x,w)}\cdot K_{(1,w)}\cdot V_{(1,y)} \notag\\
 &+   Q_{(x,1)}\cdot  K_{(2,1)}\cdot V_{(2,y)} + Q_{(x,2)}\cdot K_{(2,2)}\cdot V_{(2,y)} + \ldots + Q_{(x,w)}\cdot K_{(2,w)}\cdot V_{(2,y)} \notag\\
 &+   \vdots         	                                                                                 \notag\\
 &+   Q_{(x,1)}\cdot  K_{(h,1)}\cdot V_{(h,y)} + Q_{(x,2)}\cdot K_{(h,2)}\cdot V_{(h,y)} + \ldots + Q_{(x,w)}\cdot K_{(h,w)}\cdot V_{(h,y)} \notag
\end{align}
\begin{align}
A_{(r,c)}=\sum_{j=1}^w \sum_{i=1}^h Q_{(r,j)}K_{(i,j)}V_{(i,c)} 
\end{align}

If we joint three matrices on the dimensions with the same length, as depicted in fig\ref{fig:att2}, we will get a general area including attention matrix(A). To any unit(a) on matrix A, its value is contributed by three kinds of units in Q, K, V: with the traversal of units(k) on K, we got the other two units(q\&v) by leading a projection to row-r and column-c. It proves that unit-a collects information from every unit in K, which is different from the traditional CNN module. Again, by repetitively traversing all units a, matrix A is acquired. In conclusion, the attention used in Transformer module is repeating the collection of information from orthogonal locations.

\begin{figure}[h]
  \begin{minipage}[b]{0.3\textwidth}
	 \includegraphics[width=0.85\linewidth]{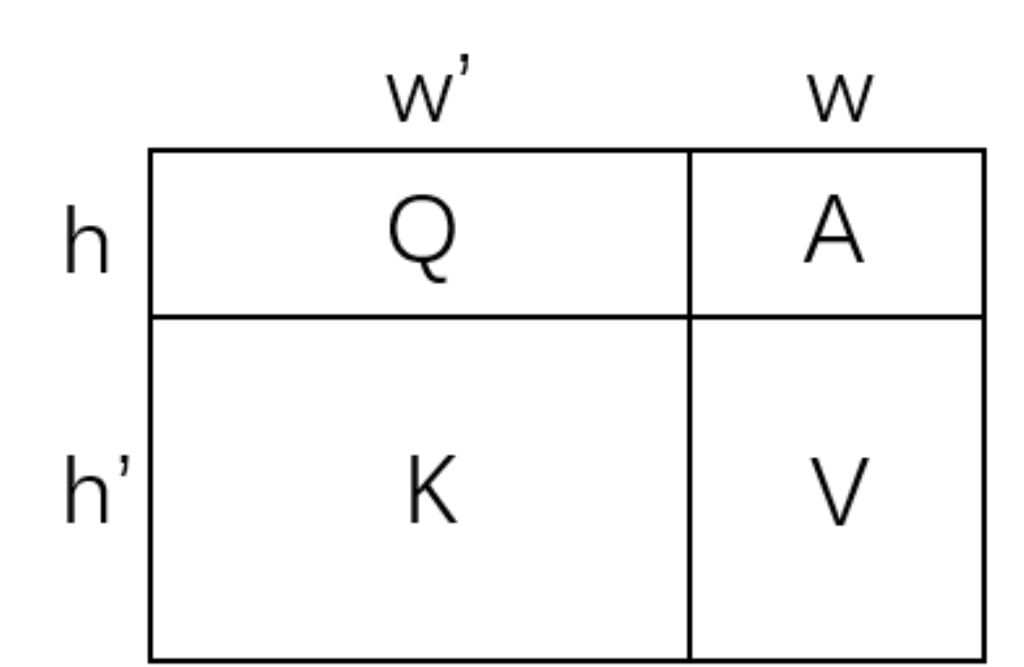}
	 \label{fig:memory1}
  \end{minipage}
  \begin{minipage}[b]{0.7\textwidth}
    \renewcommand{\arraystretch}{1.5}
    \begin{tabular}[t]{c|c|c}
      \hline
      Process & Computational Amount & Memory Occupation\\
      \hline
      $P=Q\cdot K^T$ & $w'*h*h'$ & $h*h'$\\
      $A=P\cdot V$ & $h'*h*w$ & $h*w$\\
      \hline
    \end{tabular}
	\label{tab:memory1}
    \\
  \end{minipage}
  \caption{In total: computation amount = $h*h'*(w+w')$ memory occupation = $h*(h'+w)$}
  \label{fig:simp1}
\end{figure}

\begin{figure}[h]
  \begin{minipage}[b]{0.3\textwidth}
    % \centering
    \includegraphics[width=0.8\linewidth]{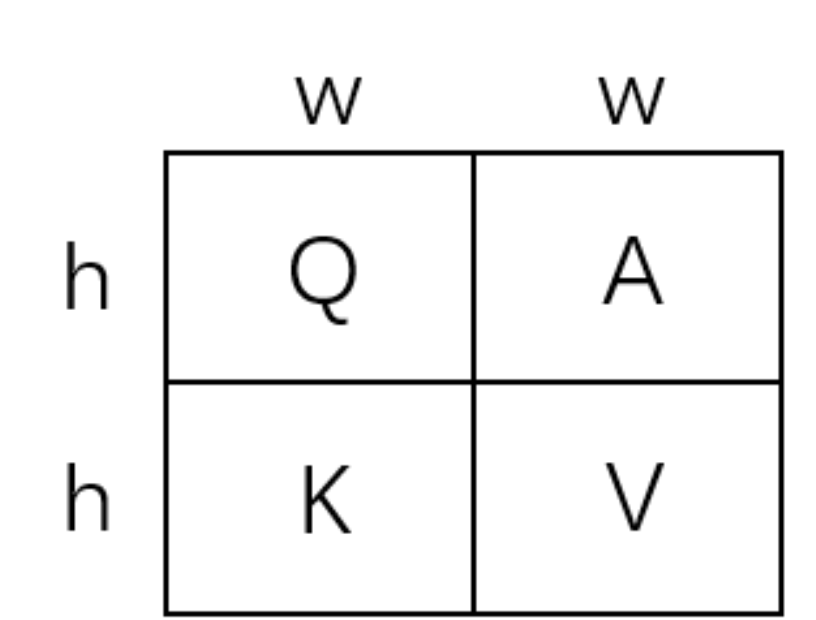}
    \label{fig:memory2}
  \end{minipage}
  \begin{minipage}[b]{0.65\textwidth}
    % \centering
    \renewcommand{\arraystretch}{1.5}
    \begin{tabular}[t]{c|c|c}
      \hline
      Process & Computational Amount & Memory Occupation\\
      \hline
      $P=Q\cdot K^T$ & $w*h*h$ & $h*h$\\
      $A=P\cdot V$ & $h*h*w$ & $h*w$\\
      \hline
    \end{tabular}
    \label{tab:memory2}
    \\
  \end{minipage}
  \caption{For self attention(h=h', w=w'),
       In total: computation amount = $2h*h*w$
       memory occupation = $h*(h+w)$}
  \label{fig:simp2}
\end{figure}

\textbf{B. Computational Amount and Memory Occupation}\\
As mentioned above, the dot-product attention has a hidden matrix P, which has a large number of intermedia data to be saved during the forward process. In fig\ref{fig:simp1}, the figure depicts the simplest QKVA grid where each matrix has various ratios and the table shows the amount of computation and memory occupation. Similarly, we make another grid that corresponds with self-attention.

\begin{figure}[h]
  \centering
   \includegraphics[width=0.4\linewidth]{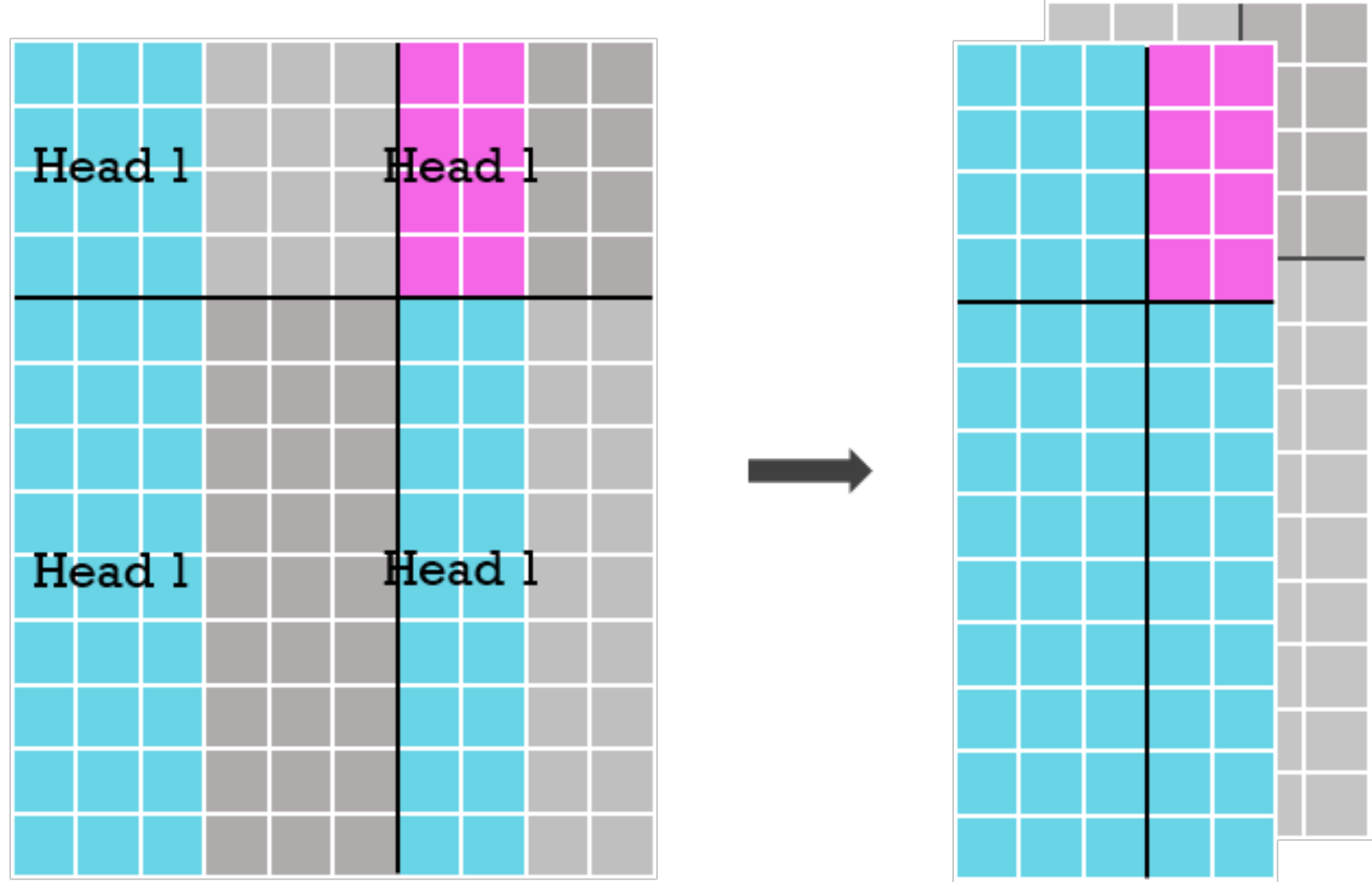}
   \caption{two-head attention in QKVA grid}
   \label{fig:att3}
\end{figure}

\textbf{C. Multi-Head}\\
Fig \ref{fig:att3} depicts the meaning of multi-head when Q, K, V share various ratios. The matrices are grouped on the feature dimension so that one head would collect information from one feature subset. Multi-head cuts down the influence between every subset.
Based on this, we find that the work of multi-head does not influence the amount of computation but makes a great impact on memory occupation. In fig \ref{fig:att3}, $w$ is the length of features. When n-head attention parts $w$ into n groups, each subset’s computation would be reduced to $1/n$, so the total computation keeps the same. While it does not change the memory occupation of each process, which makes the total memory occupation n-times larger than the single-head.

\textbf{D. Windows Attention}\\
\begin{figure}[h]
  \begin{minipage}[b]{0.7\textwidth}
    \centering
	\includegraphics[width=0.9\linewidth]{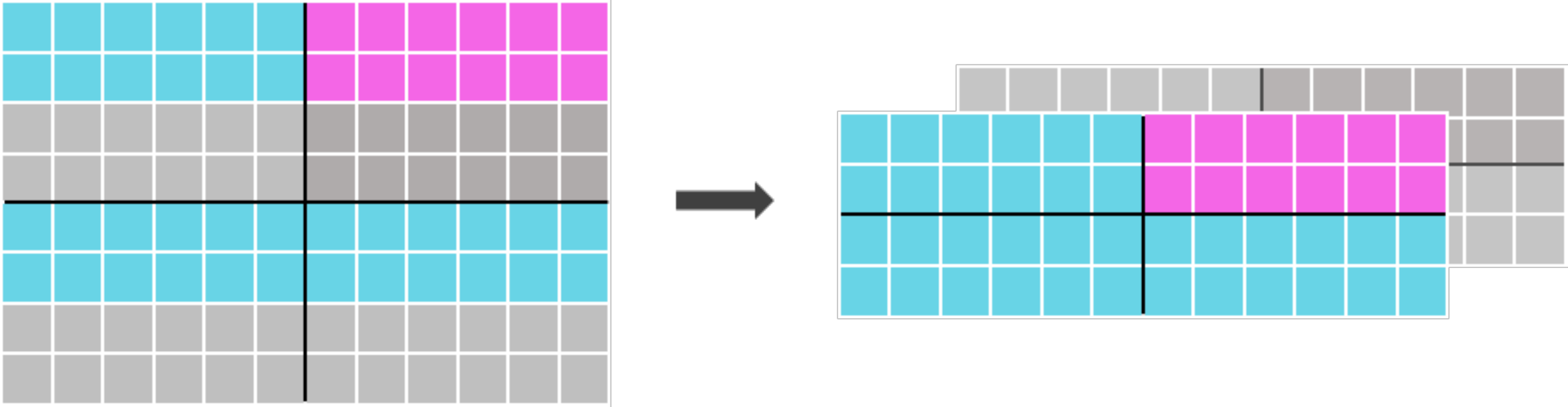}
	\subcaption{group in rows' perspevtive}
    \label{fig:att4}
  \end{minipage}
  \begin{minipage}[b]{0.25\textwidth}
    \centering
    \includegraphics[width=0.85\linewidth]{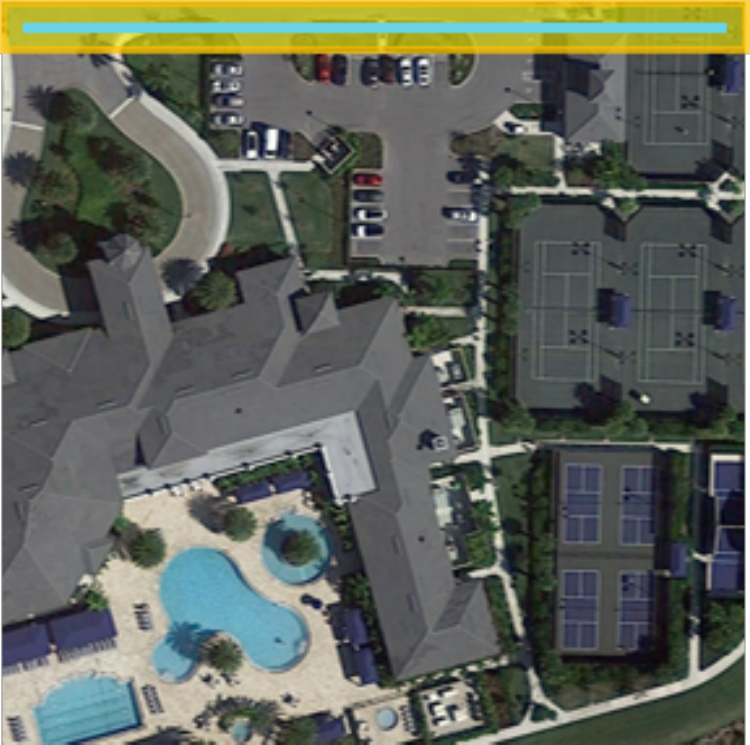}
	\subcaption{}
	\label{fig:window1}
  \end{minipage}
  \caption{}
  \label{fig:windows}
\end{figure}

\begin{figure}[h]
  \begin{minipage}[b]{0.45\textwidth}
    \centering
	\includegraphics[width=0.6\linewidth]{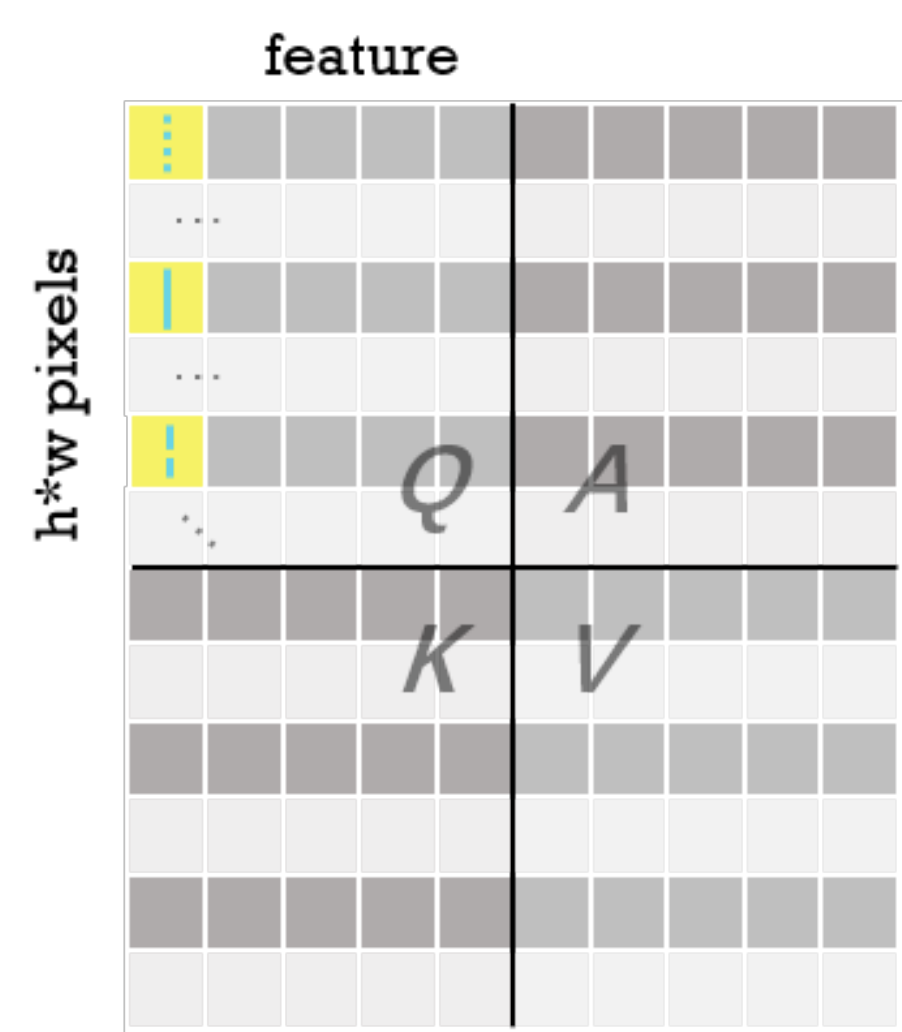}
	\subcaption{the certain pixels is described with yellow color and unique blue lines}
    \label{fig:window2}
  \end{minipage}
  \begin{minipage}[b]{0.2\textwidth}
    \centering
	\includegraphics[width=0.9\linewidth]{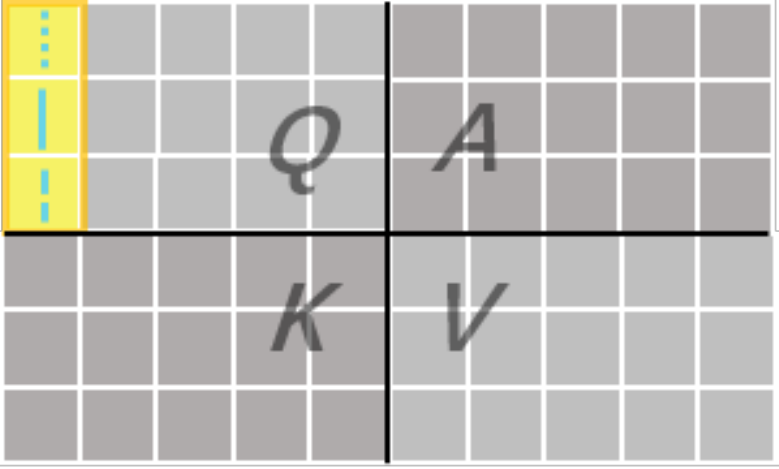}
	\subcaption{}
    \label{fig:window3}
  \end{minipage}
  \begin{minipage}[b]{0.3\textwidth}
    \centering
    \includegraphics[width=0.85\linewidth]{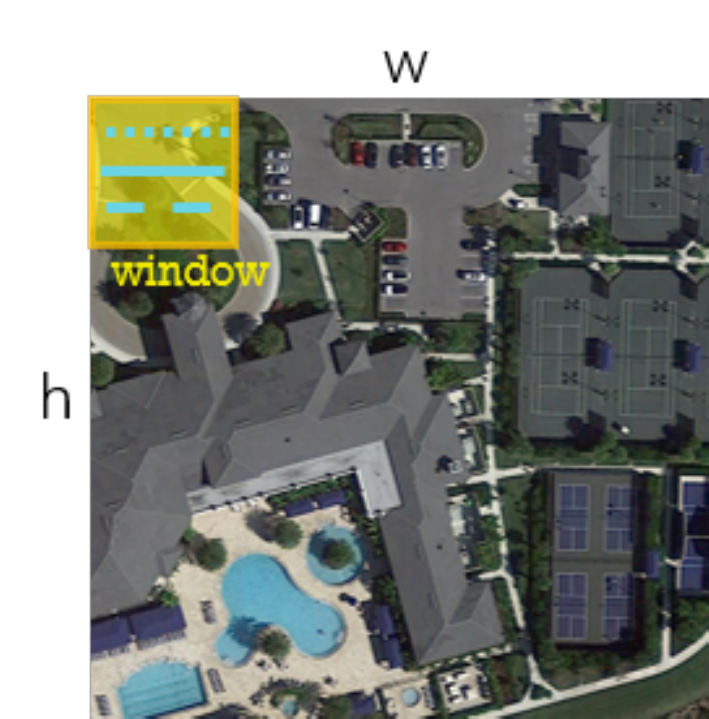}
	\subcaption{}
	\label{fig:window4}
  \end{minipage}
  \caption{}
  \label{fig:windows}
\end{figure}

In QKVA grid, multi-head groups the matrices in columns(feature dimension), and each column is isolated from the others. Also, it is possible to view the QKVA grid in rows. Fig \ref{fig:att4} looks similar as fig \ref{fig:att3} but it only works for self-attention. For the operation of input, it is popular to view the $h*w$ image as $h$ times of lines in $w$ length(we call it hw pixels). When thinking backward, the colorful subset corresponds to one row in the input image. Moreover, if we select every certain part of pixels on $w$ dimension(the yellow units in fig \ref{fig:window2}), it would correspond to a square in the image. The square is same as `window' in Swin-Transformer \cite{2021swin}, and it is the basic unit of localized attention.

\end{document}